\documentclass[letterpaper, 10 pt, conference]{ieeeconf}  

\IEEEoverridecommandlockouts                              

\usepackage{graphics} 
\usepackage{epsfig} 
\usepackage{times} 
\usepackage{amsmath} 
\usepackage{amssymb}  
\usepackage{mathtools}
\usepackage[english]{babel}

\usepackage[pagebackref,breaklinks,colorlinks]{hyperref}
\usepackage[capitalize]{cleveref}
\usepackage[dvipsnames]{xcolor}
\usepackage{nicematrix}

\usepackage{graphicx}
\usepackage{booktabs}
\usepackage{dsfont}

\usepackage{bbding}
\usepackage{pifont}
\usepackage{wasysym}
\usepackage{color, colortbl}

\usepackage{algorithm}
\usepackage{algpseudocode}
\usepackage{stackengine}
\usepackage{multirow}
\usepackage{tabularx}
\usepackage{subcaption}
\usepackage{makecell}

\title{\LARGE \bf
    Addressing Diverging Training Costs using BEVRestore \\
    for High-resolution Bird's Eye View Map Construction
}

\author{Minsu Kim$^{1\dagger}$, Giseop Kim$^2$, Sunwook Choi$^{2*}$%
\thanks{$^{1}$Minsu Kim is with Korea Institute of Science and Technology {\tt\small minsukim@kist.re.kr}}%
\thanks{$^{2}$Giseop Kim and Sunwook Choi are with vision group of NAVER LABS {\tt\small \{giseop.kim, sunwook.choi\}@naverlabs.com}}%
\thanks{$^*$Corresponding author.}
\thanks{$\dagger$ Work done during an internship at NAVER LABS.}}

\begin{document}

\maketitle
\thispagestyle{empty}
\pagestyle{empty}

\begin{abstract}
Recent advancements in Bird's Eye View (BEV) fusion for map construction have demonstrated remarkable mapping of urban environments. However, their deep and bulky architecture incurs substantial amounts of backpropagation memory and computing latency. Consequently, the problem poses an unavoidable bottleneck in constructing high-resolution (HR) BEV maps, as their large-sized features cause significant increases in costs including GPU memory consumption and computing latency, named \textit{diverging training costs} issue. Affected by the problem, most existing methods adopt low-resolution (LR) BEV and struggle to estimate the precise locations of urban scene components like road lanes, and sidewalks. As the imprecision leads to risky motion planning like collision avoidance, the diverging training costs issue has to be resolved. In this paper, we address the issue with our novel \textit{BEVRestore} mechanism. Specifically, our proposed model encodes the features of each sensor to LR BEV space and restores them to HR space to establish a memory-efficient map constructor. To this end, we introduce the BEV restoration strategy, which restores aliasing, and blocky artifacts of the up-scaled BEV features, and narrows down the width of the labels. Our extensive experiments show that the proposed mechanism provides a plug-and-play, memory-efficient pipeline, enabling an HR map construction with a broad BEV scope. Our code will be publicly released.

\end{abstract}

\vspace{15pt}\section{INTRODUCTION} \label{sec:intro}
Bird's Eye View (BEV) provides an egocentric agent with a comprehensive view of large scenes, facilitating the understanding of a global scene. Owing to its capabilities, it is widely used in constructing urban maps for autonomous driving. Recent BEV representation of deep nets applies a memory-efficient 2D space with infinite-height voxels, and associates a unified space for various modalities, such as radar, LiDAR, and camera. This approach enables the sensors to share their estimated 3D geometries and promotes collaborative sensing. For example, cameras effectively complement the sparse data distribution of radar and LiDAR with their dense sensing, while LiDAR robustly and broadly collects surround geometry, and radar can estimate accurate poses of objects. Thus, their collaborative fusion in a high-dimensional BEV space enhances a robotic system's scene understanding \cite{kim2023broadbev}.

\begin{figure}[t]
    \footnotesize
    \centering
    \begin{subfigure}[b]{\linewidth}
        \centering
        \stackunder[5pt]{\includegraphics[width=0.325\linewidth]{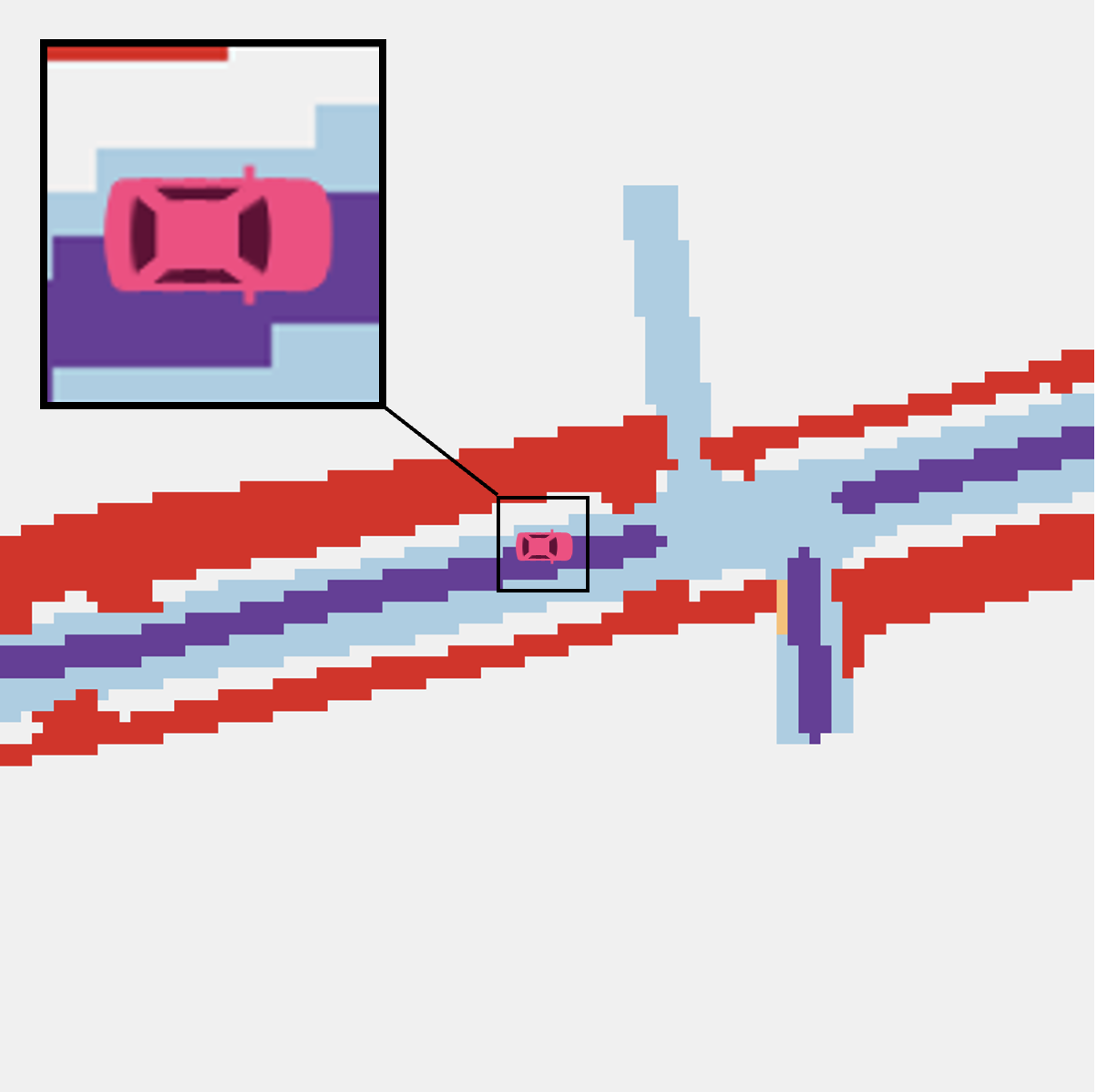}}{1.0m/px}
        \stackunder[5pt]{\includegraphics[width=0.325\linewidth]{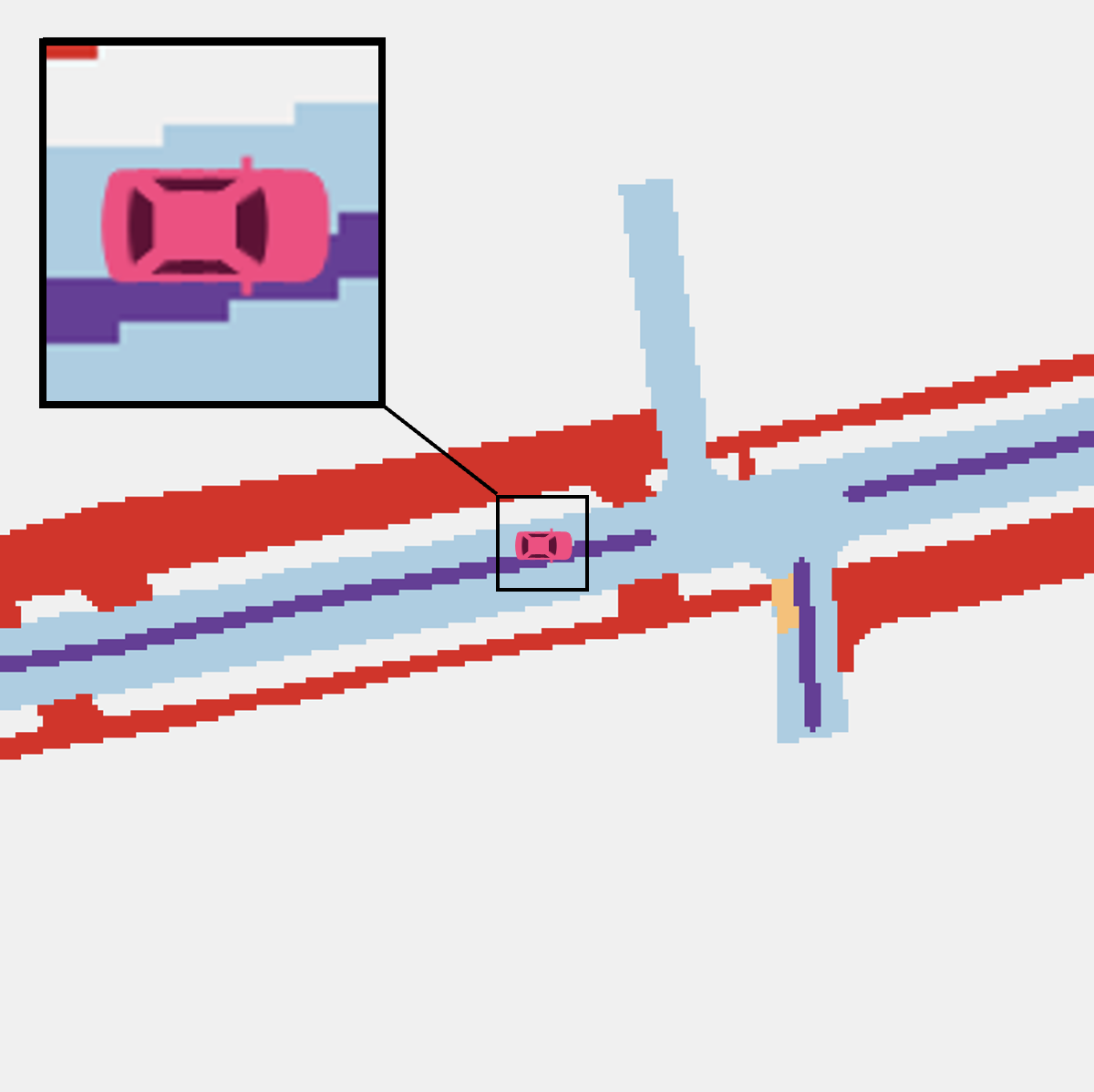}}{0.5m/px, \textbf{$\times$2 Up}}
        \stackunder[5pt]{\includegraphics[width=0.325\linewidth]{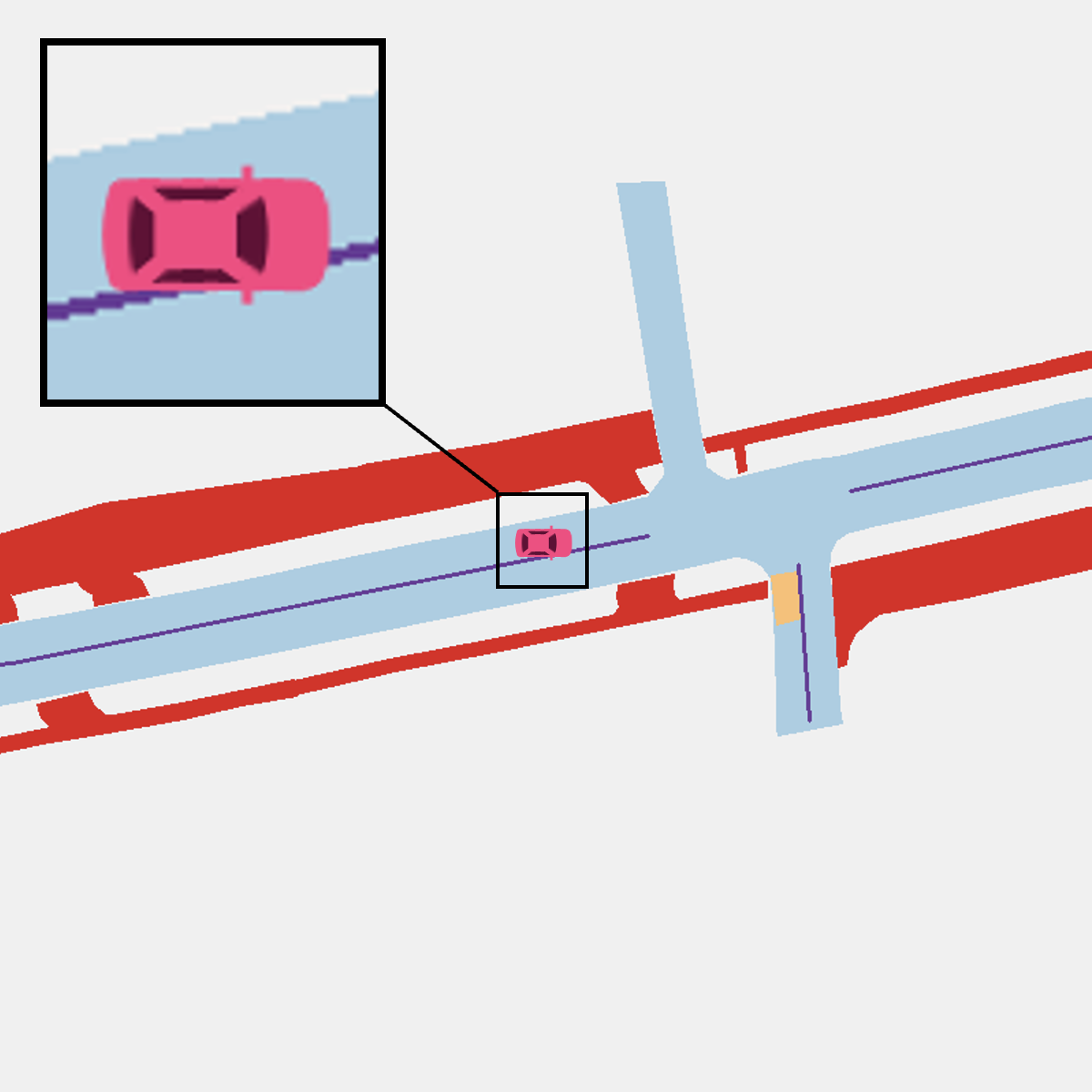}}{0.125m/px, \textbf{$\times$8 Up}}
        \subcaption{High-resolution map construction using BEVRestore.}
        \label{subfig:bevrestore}
        \vspace{4pt}
    \end{subfigure}
    \begin{subfigure}[b]{0.57\linewidth}
        \centering
        \includegraphics[width=1.05\linewidth]{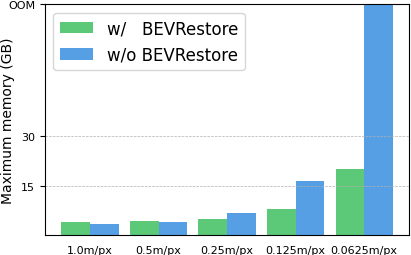}
        \subcaption{Training memory efficiency.}
        \label{subfig:cost}
    \end{subfigure}
    \hspace{2pt}
    \begin{subfigure}[b]{0.40\linewidth}
        \centering
        \includegraphics[width=0.735\linewidth]{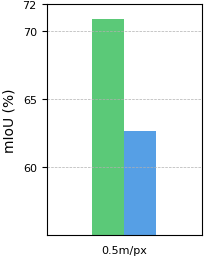}
        \subcaption{BEV Seg. performance.}
        \label{subfig:performance}
    \end{subfigure}
    \vspace{-10pt}
    \caption{``m/px" means meter per pixel. Training with HR BEV features consumes a huge memory size due to the bulky architecture. Our plug-and-play BEVRestore (\cref{subfig:bevrestore}) addresses the issue, allowing for costly efficient map construction (\cref{subfig:cost}) and enhancing feature encoding (\cref{subfig:performance}).}
    \label{fig:demo}
    \vspace{-19pt}
\end{figure}

However, existing BEV fusions face a significant hurdle in inferring high-resolution (HR) BEV representations (See \cref{subfig:bevrestore} for low and high BEV resolutions). Their heavy architecture, which includes multiple encoders, decoders, feature pyramid networks \cite{lin2017feature}, and a feature lifting \cite{philion2020lift} module, cause substantial memory consumption and computational bottlenecks. Consequently, when a BEV is configured with a higher resolution or heavy mechanism like attention \cite{vaswani2017attention}, both backpropagation memory and computing latency increase significantly. Because the issue, which we term the \textit{diverging training costs}, limits the learning of HR BEVs with high costs, recent self-driving agents rely on low-resolution (LR) BEVs exposing them to risky behaviors such as crossing stop lines or road lanes. For example, in \cref{fig:demo}, despite depicting the same section of the road, the low-resolution maps indicate that the vehicle remains within its lane, while the high-resolution maps definitively show that the vehicle crosses the lane boundary. Since this misunderstanding endangers safe motion planning, we have to address this problem. To this end, it is necessary to explore efficient pipelines for representing an HR BEV, thereby it becomes generally available in self-driving fields.

In this paper, we introduce a plug-and-play, memory-efficient approach, BEVRestore. Our method encodes the context features of each sensor into an LR BEV space. Then, after feature enhancement, BEVRestore reconstructs local details. Our approach alleviates diverging training costs while achieving performance gains establishing a new baseline for BEV map construction. In our experiments, we focus on examining the performance improvement and compatibility of BEVRestore across sensors, BEV encoders, and map construction tasks including BEV segmentation and HD map construction. Our contributions are summarized as:

\begin{itemize}
\setlength{\leftmargin}{-.35in}
    \item Our proposed mechanism addresses diverging training costs, thereby facilitating efficient training of models for representing high-resolution BEVs, which are crucial for safe self-driving. \\
    \vspace{-8pt}

    \item BEVRestore is the first framework to utilize restoration of low-resolution BEV, which up-samples BEV features, thins the semantic labels, and restores their aliasing and blocky artifacts. Owing to this costly novel approach, BEVRestore efficiently constructs HR BEV maps addressing the diverging training costs. \\
    \vspace{-8pt}

    \item Our approach improves map construction performance. We extensively investigate the property in addition to its compatibility across various sensors, BEV encoders, and map construction tasks.\\
    \vspace{-14pt}
\end{itemize}

\section{RELATED WORKS} \label{sec:related}
Fusion of LiDAR-camera BEV has shown promising demonstrations in map construction \cite{roddick2018orthographic, hendy2020fishing, hu2021fiery, zhou2022cross, li2022bevformer, yang2023bevformer, li2022hdmapnet, man2023bev, borse2023x, wang2023lidar2map, liu2023bevfusion, vora2020pointpainting, yin2021multimodal, ge2023metabev, wang2023unitr, wang2023lidar2map, kim2023broadbev, sharma2023bevseg2tp, kim2023crn, zhu2023mapprior, chang2024bevmap}. Borse \textit{et al.} suggested X-Align \cite{borse2023x} and showed that domain alignment enhances the robustness of BEV features against noisy and blurred effects. Kim \textit{et al.} \cite{kim2023broadbev} introduced a methodology for broad BEV perception. Wang \textit{et al.} \cite{wang2023unitr} suggested an effective transformer architecture for multi-modal fusion. Kim \textit{et al.} suggested CRN \cite{kim2023crn}, a camera-radar fusion utilizing prospectively projected camera images for enriching semantic camera BEV features. Chang \textit{et al.} proposed a BEVMap \cite{chang2024bevmap}, which utilizes a pre-built map to address the inaccurate placement of camera BEV features in 3D space. HDMapNet \cite{li2022hdmapnet} suggests a general framework for HD map construction with BEV segmentation. LiDAR2Map \cite{wang2023lidar2map} proposed a distillation strategy to construct HD maps with LiDAR-only modality. Shin \textit{et al.} suggested InstaGraM \cite{shin2023instagram}, a real-time HD map constructor. Qiao \textit{et al.} proposed BeMapNet \cite{qiao2023end}, a map constructor with bezier curves, and Machmap \cite{qiao2023machmap}, an end-to-end solution for BEV map construction.

Typically, the configurations of BEV scope and resolution they employ are categorized according to their task-specific aims. For example, to construct an HD map, a model should densely perceive the urban environment. Thus, most models adopt (-30m, 30m) X-axis and (-15m, 15m) Y-axis scope in nuScenes LiDAR coordinate system \cite{caesar2020nuscenes} with 0.15m/px. In contrast, BEV segmentation utilizes (-50m, 50m) X and Y-axis scope with 0.5m/px resolution allowing for sparse, thereby utilizing relatively small-sized BEV features. Thus it enables a broad perception range. 

Considering the diverging training costs of HR BEV fusion systems, there is no alternative but to use the task-specific BEV scopes as the dense and broad urban mapping requires large-sized tensor templates of BEV features. For instance, if a map constructor uses a broad and dense (-50m, 50m, 0.1m/px) scope, it would encounter severe computational overheads and memory consumption due to the size of BEV features ($1000 \times 1000\times C$). Therefore, utilizing the HR variables with limited hardware resources necessitates narrowing perception ranges to mitigate diverging training costs. To address the problem, we suggest BEVRestore, which processes coarsely voxelized BEVs and outputs an up-sampled HR BEV representation.

\section{PROBLEM FORMULATION} \label{sec:formulation}
This section outlines the formulation of BEVRestore ($\mathbf{G}_{\theta}$), where $\theta$ denotes parameter. First, we revisit the conventional representation of a fused BEV feature map. Then, we describe the BEVRestore mechanism and clarify the challenges of its implementation. Our formulations are under the assumption that a camera ($\mathbf{z}_i \in \mathbb{R}^{C_i}$), a LiDAR ($\mathbf{z}_p \in \mathbb{R}^{C_p}$), and a fused ($\mathbf{z} \in \mathbb{R}^{C_f}$) BEV feature are prepared from $K$ images of multiple cameras ($\mathbf{I}$) and $N$ points of LiDAR ($\mathbf{P}$) as in Liu \textit{et al}'s method \cite{liu2023bevfusion}. We respectively notate $D\times W$ and $d\times w$ as HR and LR sizes.

\subsection{Convention of BEV Fusion}
The existing approaches \cite{liu2023bevfusion, borse2023x, man2023bev, kim2023broadbev} extract an optimally fused BEV feature $\mathbf{z}^{*}\in \mathbb{R}^{D\times W\times C_f}$ as:
\begin{gather}
    \mathbf{z}^{*}=\arg\min_{\hat{\mathbf{z}}} \;\; CE(M, \hat{M}), \\
    \textrm{where} \;\; \hat{M} = \mathbf{D}_\phi(\hat{\mathbf{z}}), \;\; \hat{\mathbf{z}} = \mathbf{B}_\psi(f_\nu(\mathbf{z}_p,\; \mathbf{z}_i)),
\end{gather}
`$CE$' means the standard cross-entropy with focal loss \cite{lin2017focal} between ground-truth ($M\in \mathbb{R}^{D\times W}$) and predicted ($\hat{M}\in \mathbb{R}^{D\times W}$) maps. The subscript of $\mathbf{z}_{(\cdot)}$ means the data modality, where $i$ and $p$ respectively correspond to the images and points from the camera and LiDAR, respectively. $\mathbf{D}_\phi$ is a decoding head for BEV segmentation with learnable parameter $\phi$, $f_\nu(\cdot, \cdot): (\mathbb{R}^{C_p}, \mathbb{R}^{C_i}) \mapsto \mathbb{R}^{C_f}$ denotes neural networks with parameter $\nu$ for BEV fusion such as CNN, Transformer, $\mathbf{B}_\psi(\cdot): \mathbb{R}^{C_f} \mapsto \mathbb{R}^{C}$ denotes cross-modal fusion neck enhancing BEV features.

\begin{figure*}[t]
    \vspace{3mm}
    \centering
    \includegraphics[width=0.93\linewidth]{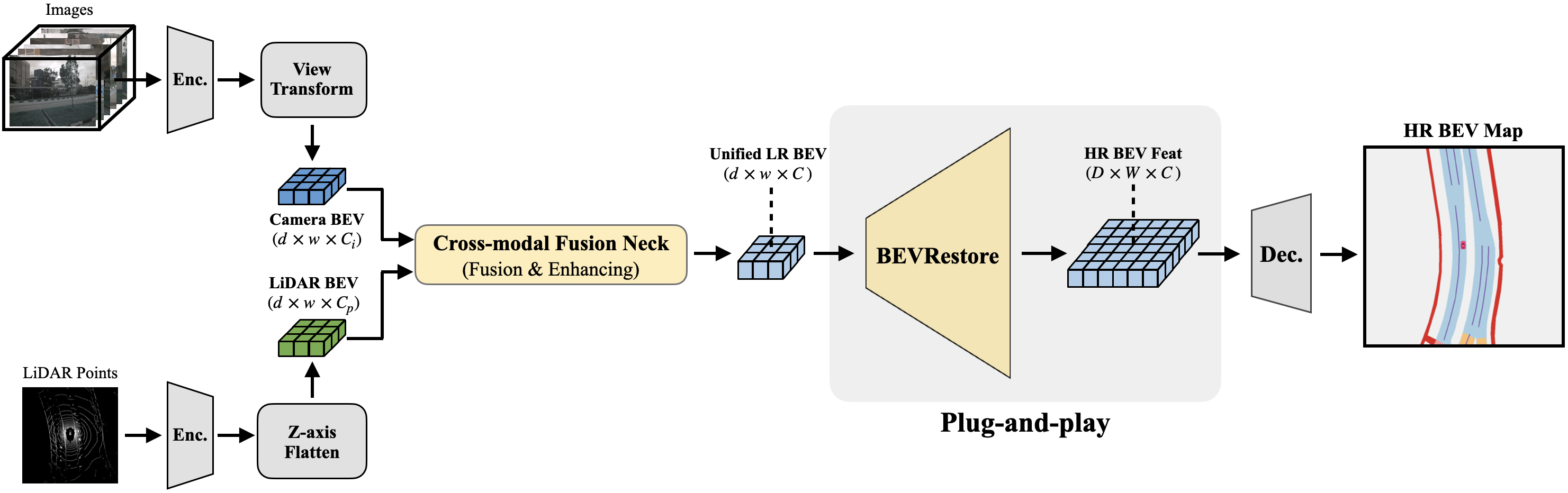}
    \caption{\textbf{Overview of Our Proposed Methods.} Our suggested mechanism take in LiDAR ($\mathbf{z}^{LR}_p$) and Camera ($\mathbf{z}^{LR}_i$) BEV features. After the cross-modal fusion neck ($\mathbf{B}_\psi$) fuse and enhance them, our BEVRestore ($S$) up-samples and restores a unified HR BEV feature. Then decoding CNNs ($\mathbf{D}_\phi$) estimate an HR semantic map.}
    \label{fig:flowchart}
    \vspace{-16pt}
\end{figure*}


\subsection{Our Strategy}
A semantic BEV map for driving perception is composed of high-level global features due to its broad scope in representing urban scenes. Our approach leverages LR space to prioritize the learning of global features over local details and restores the LR BEV to HR space. To this end, our proposed BEVRestore leverages Pixel Shuffle \cite{shi2016real} as:
\begin{spreadlines}{5pt}
\begin{gather}
    \hat{\mathbf{z}} = S(\mathbf{B}_\psi(f_\nu (\mathbf{z}^{LR}_p,\; \mathbf{z}^{LR}_i)); \varphi, s),\;\;
    S = \mathcal{PS} \circ f_\varphi, \label{eq:criteria}
\end{gather}
\end{spreadlines}
$S(\cdot; \varphi): \mathbb{R}^{d\times w} \mapsto \mathbb{R}^{D\times W}$ is a sequential operation of neural networks ($f_\varphi$) followed by Pixel Shuffle ($\mathcal{PS}$), $\mathbf{z}^{LR} \in \mathbb{R}^{C_f}$ is an LR BEV feature, $s$ is a down-sizing scale factor. With our HR BEV evaluation in \cref{eq:criteria}, $f_\varphi$ learns the restored HR BEV representation, and the composed neural nets learn compressed semantic labels as in the zoom-in regions of \cref{fig:doublebell}. In urban BEV maps, especially for roads and their surroundings, layouts are often categorized into a few cases with minimal variation. Key points like intersections (``nodes") and the roads connecting them (``edges") follow standardized patterns \cite{lynch1964image} that neural networks can easily approximate. LR BEV has sufficient representation power for this. Our approach generates semantic labels in the LR BEV space and up-samples them to HR BEV, restoring any blocky features.

\section{METHOD} \label{sec:method}
The overall pipeline is illustrated in \cref{fig:flowchart}. Our framework takes in $K$ images and $N$ LiDAR points to estimate a semantic BEV map. It adheres to the conventions established by BEVFusion \cite{liu2023bevfusion} for extracting camera and LiDAR BEV features. Subsequently, the BEVRestore unifies the two BEV features into a unified BEV feature map and enhances it with a fusion neck. Afterward, it deploys BEVRestore to construct HR BEV and decode them into a semantic map.

\subsection{Evaluation of Low-resolution BEV}
In our equations, the range of BEV grid $\mathbf{R}=\textrm{($lb$, $ub$, $r$)}$ represents `$lb$' as the lower and `$ub$' as the upper bound distances from an egocentric vehicle with `$r$' meters per pixel (m/px) resolution. Each trained model assumes that $(lb, ub, r)$ is fixed for all the BEV representation axes.

\noindent\textbf{Camera Branch.} When given backbone nets ($\mathbf{E}_{\omega, i}$) and a down-sizing scale factor ($s$), the BEV features ($\mathbf{z}_i \in \mathbb{R}^{H/s \times W/s \times C_i}$) are evaluated as:
\begin{gather}
    \mathbf{z}_i = \mathbf{T}(\mathbf{E}_{\omega, i}(\mathbf{I}), \mathbf{R}; s), \label{eq:cam}
\end{gather}
where $\mathbf{T}$ denote view transform \cite{liu2023bevfusion}. The learnable parameters are predominantly located in backbone networks ($\mathbf{E}_{\omega, i}$), and as described in \cref{eq:cam}, the encoding of image features is not hindered by a shortage of local details. Thus, the extraction of camera BEV features prevents the degradation of representational capacity for generalizable perception.

\noindent\textbf{LiDAR Branch.} After $N$ LiDAR points ($\mathbf{P}=\{p_1, p_2, ..., p_N\}$) are voxelized, their features within a common X and Y BEV range ($\mathbf{R}$), are extracted. When given backbone nets ($\mathbf{E}_{\omega, p}$), the evaluation of LiDAR BEV features ($\mathbf{z}_p \in \mathbb{R}^{H/s \times W/s \times C_p}$) is as:
\begin{gather}
    \mathbf{z}_p = \textrm{Flatten} \circ \mathbf{E}_{\omega, p}\circ\textrm{Voxelize}(\mathbf{P}, \mathbf{R}; s),
\end{gather}
where `Flatten' means the Z-axis flattening of voxelized features \cite{liu2023bevfusion}. The LR BEV offers memory-efficient computation and global features. Furthermore, LR BEV space is effectively utilized to represent the vast scope of the standardized urban scenes, as its global representation excels in approximating them.

\subsection{BEV Restoration}
Restoration of BEV simultaneously up-samples and estimates local details of HR BEV representation ($\hat{\mathbf{z}}^{HR}$) with our proposed operator ($S(\cdot; \varphi, s)$) as:
\begin{gather}
    \hat{\mathbf{z}} = S(\hat{\mathbf{z}}^{LR}; \varphi, s).
\end{gather} \vspace{-18pt}

\subsection{Manipulation of BEV Scope and Resolution} \label{sec:processing}
\noindent\noindent\textbf{Spatial Pooling.} Pooling with kernels, such as a down-sampling CNN layer, max, and average poolings, enlarges the pixel-wise receptive field size and the corresponding coverage of BEV ranges. Given that $\mathbf{u} = [u, v]$ is a pixel coordinate, $\chi=\{[i, j] \;|\; i \in \chi_u,\; j \in \chi_v\}$ is a relative coordinate set for a pooling kernel with $|\chi_u|\times |\chi_v|$ size, the pre-defined BEV resolution $\mathbf{r}=[r_i, r_j]$ and the coverage of a BEV pixel $\mathcal{R}(\mathbf{u; r})$ are re-defined $(\cdot)'$ as:
\begin{gather}
    \mathcal{R}'(\mathbf{u;\; r'}) = \bigcup_{[i,j]\in \chi} \{\mathcal{R}(\mathbf{u} + [i, j];\; \mathbf{r}')\},
    \label{eq:maxpool} \\
    \textrm{where}\; \mathbf{r}' = [\;r_i \cdot |\chi_u|,\; r_j \cdot |\chi_v|\;].
\end{gather}
As in \cref{eq:maxpool}, the BEV feature and its coverage are respectively down-sampled and enlarged while becoming a global representation. 


\begin{table}[t]
    \scriptsize
    \centering
    \vspace{5pt}
    \caption{\textbf{Plug-and-play performance gains for BEV segmentation of various modalities.} We plug ``$\times 4$" BEVRestore into the camera, LiDAR, and LiDAR-camera fusion modalities, employing three distinct backbones to investigate BEVRestore's compatibility.}
    \setlength{\tabcolsep}{1.5pt}
    \renewcommand{\arraystretch}{1.35}
    \begin{subtable}[t]{\linewidth}
        \scriptsize
        \centering
        \vspace{-5pt}
        \begin{tabular}{
            c
            |>{\centering\arraybackslash}p{0.30\columnwidth}
            |>{\centering\arraybackslash}p{0.18\columnwidth}
            |>{\centering\arraybackslash}p{0.18\columnwidth}
        }
            \Xhline{3\arrayrulewidth}
            \multirow{2}{*}{Method} & \multirow{2}{*}{Modality} & \multirow{2}{*}{$\times 4$} & \multirow{2}{*}{mIoU} \\
             & & & \\
            \Xhline{3\arrayrulewidth}
            \multirow{2}{*}{Swin-T \cite{liu2021swin} + LSS \cite{philion2020lift}} & \multirow{2}{*}{Camera} &  & 57.1 \\
             &  & \ding{51} & \textbf{61.4} \\
            \Xhline{0.1\arrayrulewidth}
            \multirow{2}{*}{Sparse Conv. \cite{liu2015sparse}} & \multirow{2}{*}{LiDAR} &  & 48.6 \\
             &  & \ding{51} & \textbf{68.7} \\
            \Xhline{0.1\arrayrulewidth}
            \multirow{2}{*}{BEVFusion \cite{liu2023bevfusion}} & \multirow{2}{*}{LiDAR-camera Fusion} &  & 62.7 \\
              &  & \ding{51} & \textbf{70.9} \\
            \Xhline{3\arrayrulewidth}
        \end{tabular}
    \end{subtable}
    \label{tab:Quan_modality}
    \vspace{-2pt}
\end{table}

\begin{table}[t]
    \setlength{\tabcolsep}{1.5pt}
    \renewcommand{\arraystretch}{1.44}
    \caption{\textbf{Plug-and-play performance gains for HD map construction with camera BEV.}  We plug ``$\times 4$" BEVRestore into HD map construction models.}
    \label{tab:other_baseline}
    \begin{subtable}[t]{\linewidth}
        \scriptsize
        \centering
        \label{tab:plug2}
    \begin{tabular}{c
        |>{\centering\arraybackslash}p{0.11\textwidth}
        |>{\centering\arraybackslash}p{0.13\textwidth}
        >{\centering\arraybackslash}p{0.13\textwidth}
        |>{\centering\arraybackslash}p{0.17\textwidth}
        >{\centering\arraybackslash}p{0.17\textwidth}
    }
        \Xhline{3\arrayrulewidth}
        \multirow{2}{*}{Method} & \multirow{2}{*}{$\times 4$} & \multirow{2}{*}{mAP} & \multirow{2}{*}{mIoU} & Latency & Mem. \\
         & & & & (ms) & (GB) \\
        \Xhline{3\arrayrulewidth}
        \multirow{2}{*}{HDMapNet \cite{li2022hdmapnet}} & & 23.0 & 34.2 & \textbf{38} & \textbf{1.9} \\
         & \ding{51} & \textbf{32.1} & \textbf{36.4} & 58 & 3.5  \\
        \Xhline{3\arrayrulewidth}
        \multirow{2}{*}{BEVFormer \cite{li2022bevformer}} &  & - & 40.1 & 125 & 9.9 \\
        & \ding {51} & - & \textbf{41.3} & \textbf{104} & \textbf{2.5} \\
        \Xhline{3\arrayrulewidth}
    \end{tabular}
    \end{subtable}
    \vspace{-15pt}
\end{table}

\noindent\textbf{BEVRestore}. The operation consists of pixel shuffle ($\mathcal{PS}$) and CNNs ($f_\varphi$). In analytical point of view, $f_\varphi(\cdot): \mathbb{R}^{h\times w\times C} \mapsto \mathbb{R}^{h\times w\times s^2C}$ projects latent geometry features with $s$-scaled scope in their channels, $\mathcal{PS}(\cdot): \mathbb{R}^{sh \times sw\times C}$ reshapes them to an HR BEV. Thus, BEVRestore preserves BEV scope ($\mathbf{R}$) but manipulates resolution ($r$). It estimates interpolated BEV features from LR BEVs restoring local details, due to the optimization policy that imposes the model to align with the HR semantic BEV maps' ground truth.





\subsection{Implementation Details} \label{subsec:impl}
Our implementations are built on top of mmdetections \cite{chen2019mmdetection}, \cite{mmdet3d2020}. We use the Swin-T (Tiny) \cite{liu2021swin} model and feature lifting module \cite{philion2020lift} to encode the camera BEV. We adopt sparse convolution \cite{liu2015sparse} as a LiDAR backbone. To fuse the LiDAR and the camera BEV features, we use a fully convolutional layer with Feature Pyramid Networks (FPN) \cite{lin2017feature}. The data augmentations we used are the same methods as the baseline work \cite{liu2023bevfusion}. Detailed model and training configurations are shown in our publicly available code.

\begin{table}[t]
    \scriptsize
    \vspace{5pt}
    \caption{\textbf{Comparison of LR to HR BEV segmentations.} \textbf{``mIoU ($r$)"} denotes the evaluated \textbf{mIoU} of ``$r$" BEV resolution. ``BEVRestore" and ``BEVRestore-S (Small)" are pre-trained on 2.0m/px, and fine-tuned on 0.5m/px BEV space. To compose ``BEVRestore-S", We downsize the number of parameters of BEVRestore and decoding nets for pure exploration of the effectiveness of LR BEV space}
    \begin{subtable}[t]{\linewidth}
        \centering
        \renewcommand{\arraystretch}{1.2}
        \begin{tabular}{
            c|
            >{\centering\arraybackslash}p{0.20\linewidth}|
            >{\centering\arraybackslash}p{0.11\linewidth}
            >{\centering\arraybackslash}p{0.11\linewidth}
            >{\centering\arraybackslash}p{0.10\linewidth}
        }
            \Xhline{3\arrayrulewidth}
            \multirow{2}{*}{Method} & Trained & mIoU & mIoU & Param.\\
             & BEV Resolution & (2.0m/px) & (0.5m/px) & (M) \\
            \Xhline{3\arrayrulewidth}
            LR BEV & 2.0m/px & 70.8 & - & 46.8 \\
            HR BEV & 0.5m/px & - & 62.7 & 46.8 \\
            \rowcolor[gray]{0.9} $\times 4$ BEVRestore-S & 2.0 → 0.5m/px & 70.8 & 64.7 & \textbf{44.5} \\            \rowcolor[gray]{0.9} $\times 4$ BEVRestore & 2.0 → 0.5m/px & 70.8 & \textbf{70.9} & 148.3 \\
            \Xhline{3\arrayrulewidth}
        \end{tabular}
    \end{subtable}
    \vspace{-4pt}
    \label{tab:discuss}
\end{table}

\subsection{Training}
Our optimization aims to implement a modular BEVRestore mechanism. We first train a base model, which utilizes LR BEV. Afterward, we freeze the parameters before BEVRestore and fine-tune BEVRestore and decoder to use HR BEV space. This strategy enables attaching varying-scaled learnable up-sampling to the latter of the model. Furthermore, the designs address the concern that high-resolution BEV consumes large memory space generating partial backpropagation memory.



\section{EXPERIMENTAL RESULTS} \label{sec:exp}

\noindent\textbf{Dataset.} We use nuScenes \cite{caesar2020nuscenes}. The input images are resized to $256\times704$ resolution, and LiDAR points are voxelized to (0.125m, 0.125m, 0.2m), (0.125m, 0.125m, 0.2m), (0.25m, 0.25m, 0.2m), and (0.5m, 0.5m, 0.2m) XYZ resolutions for $\times2$, $\times4$, and $\times8$ scale factors, respectively.


\noindent\textbf{Ground truth.}
The semantic BEV maps of nuScenes are depicted through vectorized nodes (or geometric layers). Because the rasterization of BEV maps is derived from the nodes, any loss of ground truth is limited to negligible errors in node positions, without compromising the quality of the rasterized maps. We leverage the nuScenes map expansion \cite{nuscenes2019} to generate scalable ground truth.

\subsection{Quantitative Results} \label{subsec:quan}
\noindent\textbf{Setup.} We apply a $\times4$ BEVRestore, (-50m, 50m) BEV scope, 2.0m/px LR and 0.5m/px HR BEV resolutions in \cref{tab:Quan_modality}, \cref{tab:discuss}, \cref{tab:Quan_sr}, and \cref{tab:footprint}. In \cref{tab:other_baseline}, we use (-30m, 30m) and (-15m, 15m) BEV scopes with 0.6m/px LR and 0.15m/px HR BEV resolutions. Gray shading in each table represents our base model.

\noindent\textbf{Cost Measurement.} We examine the inference time latency and the training time maximum memory based on a single batch using an NVIDIA A100 80GB GPU.

\begin{table}[t]
    \scriptsize
    \caption{\textbf{Comparisons of up-sampling methods and computational efficiency using $\times 4$ models.} The Learnable Deconvolution and Pixel Shuffle show effective restoration. Deconvolution is an over-determined form of Pixel Shuffle that consumes large computational costs without improving performance, which is why we utilize Pixel Shuffle.}
    \centering
    \setlength{\tabcolsep}{1.5pt}
    \renewcommand{\arraystretch}{1.3}
    \begin{tabular}{
        c|
        >{\centering\arraybackslash}p{0.26\columnwidth}
        >{\centering\arraybackslash}p{0.26\columnwidth}
        >{\centering\arraybackslash}p{0.26\columnwidth}
    }
        \Xhline{3\arrayrulewidth}
        Method & Latency (ms) & Mem. (GB) & mIoU \\
        \Xhline{3\arrayrulewidth}
        Nearest & \textbf{151} & \textbf{3.6} & 62.9 \\
        Bilinear & 155 & \textbf{3.6} & 63.5 \\
        Bicubic & 158 & \textbf{3.6} & 63.6 \\
        Deconvolution & 167 & 3.9 & \textbf{70.9}\\
        \rowcolor[gray]{0.9} Pixel Shuffle & 164 & 3.8 & \textbf{70.9} \\
        \Xhline{3\arrayrulewidth}
    \end{tabular}
    \label{tab:Quan_sr}
    \vspace{-13pt}
\end{table}

\begin{table}[t]
    \centering
    \scriptsize
    \vspace{5pt}
    \caption{\textbf{Effectiveness tendency with increasing up-scale factors.} Smaller LR BEV space is effective in learning the context of huge scenes. However, excessively down-sized BEV space limits the improvement as indicated by the mIoU gap between $\times 4$ and $\times 8$ scales.}
    \begin{subtable}[t]{\linewidth}
        \centering
        \renewcommand{\arraystretch}{1.3}
        \begin{tabular}{
            c
            |>{\centering\arraybackslash}p{0.18\linewidth}
            |>{\centering\arraybackslash}p{0.05\linewidth}            >{\centering\arraybackslash}p{0.08\linewidth}
            >{\centering\arraybackslash}p{0.08\linewidth}
            >{\centering\arraybackslash}p{0.08\linewidth}
        }
            \Xhline{3\arrayrulewidth}
            \multirow{2}{*}{Method} & \multirow{2}{*}{Resolution} & \multirow{2}{*}{mIoU} & Latency & Mem. & Param. \\
             & & & (ms) & (GB) & (M) \\
            \Xhline{3\arrayrulewidth}
            Baseline & 0.5m/px & 62.7 & 176 & 3.6 & 46.8 \\
            $\times 4$ BEVRestore-S & 2.0 $\rightarrow$ 0.5m/px & 64.7 & \textbf{157} & \textbf{2.7} & \textbf{44.5} 
            \\
            \hline
            $\times 1$ BEVRestore & 0.5 $\rightarrow$ 0.5m/px & 66.3 & 218 & 5.1 & 129.4 \\
            $\times 2$ BEVRestore & 1.0 $\rightarrow$ 0.5m/px & 69.6 & 187 & 4.2 & 138.9 \\
            \rowcolor[gray]{0.9} $\times 4$ BEVRestore & 2.0 $\rightarrow$ 0.5m/px & \textbf{70.9} & 164 & 3.8 & 148.3 \\
            $\times8$ BEVRestore & 4.0 $\rightarrow$ 0.5m/px & 70.7 & \textbf{153} & 3.7 & 157.7 \\
            \Xhline{3\arrayrulewidth}
        \end{tabular}
    \end{subtable}
    \vspace{-10pt}
    \label{tab:ablation}
\end{table}

\noindent\textbf{Compatibility.} In \cref{tab:Quan_modality}, we evaluate the compatibility of the BEVRestore mechanism across various modalities, including LiDAR, camera, and LiDAR-camera fusion. As indicated in the tables, all modalities exhibit performance gains, which can be attributed to the superior representation power of LR BEV for semantic map construction, coupled with BEVRestore to evaluate HR BEV. In \cref{tab:other_baseline}, we plug BEVRestore to HDMapNet \cite{li2022hdmapnet} and a query-based camera BEV encoder BEVFormer \cite{li2022bevformer} to explore BEVRestore's compatibility for the task of vectorized semantic map construction. We follow the evaluation protocol of HDMapNet \cite{li2022hdmapnet} measuring mAP under [0.2, 0.5, 1.0] chamfer distance thresholds, and mIoU of directly estimated semantic maps. As shown in the table, our mechanism improves the map construction showing 9.1\% mAP and 2.2\% mIoU gains for vectorized and rasterized map construction of HDMapNet, 1.2\% for rasterized construction of BEVFormer. Furthermore, we find that the memory consumption of BEVFormer is significantly reduced owing to the BEVRestore mechanism's utilization of LR BEV space. However, for a light model without resource-intensive modules, BEVRestore may cause an overall increase in costs as shown in the HDMapNet's latency and memory consumption as our mechanism imposes additional costs for restoration. Despite the inevitable amount of overhead, BEVRestore prevents cost divergence achieving performance gains.

\noindent\textbf{BEVRestore.} In \cref{tab:discuss}, we analyze the restoration performance. For this purpose, we establish a baseline (or BEVFusion) trained under two different conditions: a 0.5m/px \textbf{HR BEV} space and a 2.0m/px \textbf{LR BEV} space. Additionally, we include a base model equipped with our proposed method for comparison. \textbf{BEVRestore} and \textbf{BEVRestore-S (Small)} use 2.0m/px for LR BEV and up-samples it to 0.5m/px BEV resolution. We composed BEVRestore-S to explore LR BEV space's effectiveness under the smaller representation power by downsizing the number of parameters of BEVRestore and decoding nets. As shown in the first and second rows of the table, compared to 0.5m/px resolution, BEV map construction with 2.0m/px LR BEV offers superior perception for understanding huge urban scenes owing to the LR space's condensed learning of fine details, allowing the model to focus on broader patterns and overall structures \cite{lazebnik2006beyond}. As in the third and fourth rows, Our mechanism leverages this property by locally restoring the up-sampled BEV representation. In \cref{tab:Quan_sr}, we compare trainable up-samplings with the hand-crafted approaches: `Nearest' neighbor, `Bilinear', and `Bicubic' interpolations, to validate the importance of the learnable BEVRestore. The comparisons reveal three insights. 1) Learnable up-samplings (Pixel shuffle, deconvolution) show superior BEV representation compared to handcrafted methods (Nearest neighbor, Bilinear, and Bicubic interpolations). This suggests that non-linear aliasing and blocky artifacts in BEV can be effectively restored using neural networks. 2) Deconvolution is an over-determined form of pixel shuffle as analyzed by Shi \textit{et al.} \cite{shi2016deconvolution}. Both methods yield similar performance although pixel shuffle is more cost-effective. 3) The BEVRestore mechanism's learnable up-sampling approach offers a reasonable trade-off between performance and costs. Compared to non-learnable methods, it achieves 14.2\% mIoU gain with an additional 0.4GB memory consumption and a 6ms latency increase in comparison to pixel shuffle over Bicubic interpolation.


\begin{figure}[t]
    \centering
    \vspace{5pt}
    \begin{subfigure}[b]{\linewidth}
        \centering
        \includegraphics[width=0.88\linewidth]{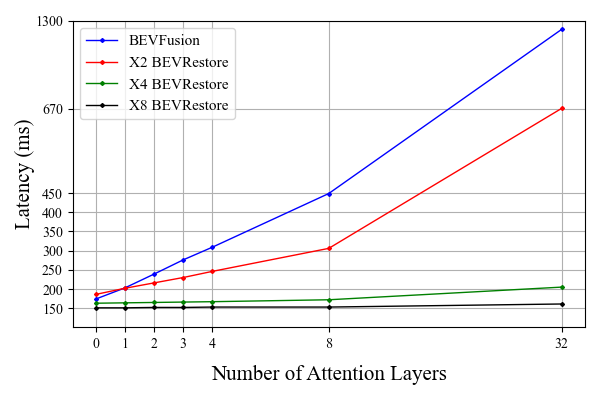}
        \vspace{-7pt}
        \subcaption{\textbf{Inference latency.}}
        \label{subfig:latency}
    \end{subfigure}
    \begin{subfigure}[b]{\linewidth}
        \centering \includegraphics[width=0.88\linewidth]{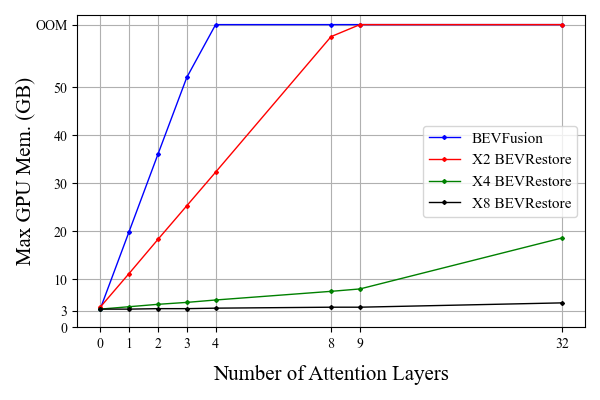}
        \vspace{-6pt}
        \subcaption{\textbf{Training GPU memory.}}
        \label{subfig:memory}
    \end{subfigure}
    \caption{\textbf{Comparison on increasing costs to BEVFusion.}} We incrementally add an MSA layer to the fusion neck and compare their costs. X and Y axes denote costs and the number of MSA layers, respectively.
    \vspace{-20pt}
    \label{fig:cost}
\end{figure}

\begin{figure*}[t]
    \vspace{3mm}
    \centering
    \begin{subfigure}[b]{\linewidth}
        \centering
        \includegraphics[width=0.94\linewidth]{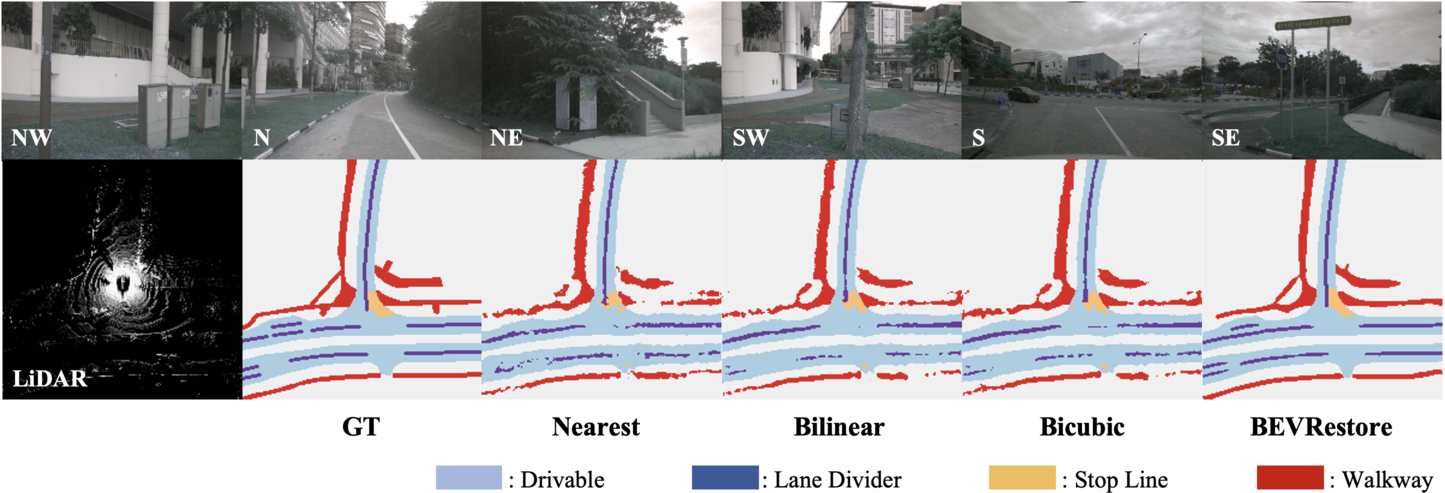}
    \end{subfigure}
    \caption{\textbf{Comparison of BEVRestore Restoration to conventional methods.} We use a (-50m, 50m) BEV scope with 2.0m/px LR BEV and 0.5m/px HR BEV resolutions. The hand-crafted \textbf{Nearest}, \textbf{Bilinear}, and \textbf{Bicubic} methods suffer from the incomplete restoration of aliasing and blocky artifacts, compared to the \textbf{BEVRestore}, which utilizes learnable restoration.}
    \label{fig:qual}
    \vspace{-18pt}
\end{figure*}

\noindent\textbf{Representation Power.}
To delve into a trade-off between efficiency and representation power, we investigate the map construction performances of varying-scaled BEVRestore in \cref{tab:ablation} with our LiDAR-camera fusion baseline, BEVFusion \cite{liu2023bevfusion}. The exploration assesses mIoU, computation latency, and maximum GPU memory consumption. We restore 0.5m/px, 1.0m/px, 2.0m/px, and 4.0m/px LR BEVs to 0.5m/px for fair comparison in (50m, 50m, 0.5m/px) BEV space. As in the table, the results imply there is a reasonable scale factor for a powerful representation of BEV with proper memory consumption as the results of the $\times 4$ scale. Although our mechanism requires additional memory to restore BEV, as shown in \cref{fig:cost}, it defenses any models across sensor modalities and perception tasks from diverging
training costs caused by computationally intensive modules, such as deep nets with increased parameters and attention layers \cite{vaswani2017attention}.

\begin{table}[t]
    \scriptsize{
    \centering
    \vspace{4pt}
    \caption{\textbf{Comparisons on cost footprints.}}
    \vspace{-2pt}
    \label{tab:footprint}
    \begin{subtable}{\columnwidth}
            \renewcommand{\arraystretch}{1.0}
            \subcaption{\textbf{(50.0m, 50.0m, 0.5m/px), BEVFusion \cite{liu2023bevfusion}.}}
            \vspace{-6pt}
            \centering
            \begin{tabular}{c
                >{\centering\arraybackslash}p{0.06\textwidth}
                >{\centering\arraybackslash}p{0.08\textwidth}
                >{\centering\arraybackslash}p{0.08\textwidth}
                >{\centering\arraybackslash}p{0.08\textwidth}
                |>{\centering\arraybackslash}p{0.08\textwidth}
                >{\centering\arraybackslash}p{0.08\textwidth}
            }
            \hline
            \multicolumn{5}{c|}{Measure Coverage} & Mem. & Latency \\
            BEV Encoder & Neck & - & Decoder & Cache & (GB) & (ms) \\
            \hline
            \ding{51} &  &  &  & & 2.3 & 158 \\
            \ding{51} & \ding{51} &  &  & & 2.4 & 167 \\
            \ding{51} & \ding{51} &  & \ding{51} & & 2.6 & 176 \\
            \ding{51} & \ding{51} &  & \ding{51} & \ding{51} & 3.6 & 176 \\
            \hline
            \end{tabular}
            \vspace{4pt}
        \end{subtable}
        \begin{subtable}{\columnwidth}
            \renewcommand{\arraystretch}{1.0}
            \subcaption{\textbf{(50.0m, 50.0m, 0.5m/px), Enhancement.}}
            \vspace{-6pt}
            \centering
            \begin{tabular}{c
                >{\centering\arraybackslash}p{0.06\textwidth}
                >{\centering\arraybackslash}p{0.08\textwidth}
                >{\centering\arraybackslash}p{0.08\textwidth}
                >{\centering\arraybackslash}p{0.08\textwidth}
                |>{\centering\arraybackslash}p{0.08\textwidth}
                >{\centering\arraybackslash}p{0.08\textwidth}
            }
            \hline
            \multicolumn{5}{c|}{Measure Coverage} & Mem. & Latency \\
            BEV Encoder & Neck & Restore & Decoder & Cache & (GB) & (ms) \\
            \hline
            \ding{51} &  &  &  & & 2.3 & 150 \\
            \ding{51} & \ding{51} &  &  & & 2.4 & 161 \\
            \ding{51} & \ding{51} & \ding{51} &  & & 4.0 & 207 \\
            \ding{51} & \ding{51} & \ding{51} & \ding{51} & & 4.1 & 218 \\
            \ding{51} & \ding{51} & \ding{51} & \ding{51} & \ding{51} & 5.1 & 218 \\
            \hline
            \end{tabular}
            \vspace{4pt}
        \end{subtable}
        \begin{subtable}{\columnwidth}
            \renewcommand{\arraystretch}{1.0}
            \subcaption{\textbf{(50.0m, 50.0m, 1.0m/px $\rightarrow$ 0.5m/px).}}
            \centering
            \vspace{-6pt}
            \begin{tabular}{c
                >{\centering\arraybackslash}p{0.06\textwidth}
                >{\centering\arraybackslash}p{0.08\textwidth}
                >{\centering\arraybackslash}p{0.08\textwidth}
                >{\centering\arraybackslash}p{0.08\textwidth}
                |>{\centering\arraybackslash}p{0.08\textwidth}
                >{\centering\arraybackslash}p{0.08\textwidth}
            }
            \hline
            \multicolumn{5}{c|}{Measure Coverage} & Mem. & Latency \\
            BEV Encoder & Neck & Restore & Decoder & Cache & (GB) & (ms) \\
            \hline
            \ding{51} &  &  &  & & 2.2 & 146 \\
            \ding{51} & \ding{51} &  &  & & 2.3 & 155 \\
            \ding{51} & \ding{51} & \ding{51} &  & & 2.9 & 176 \\
            \ding{51} & \ding{51} & \ding{51} & \ding{51} & & 3.1 & 187 \\
            \ding{51} & \ding{51} & \ding{51} & \ding{51} & \ding{51} & 4.2 & 187 \\
            \hline
            \end{tabular}
            \vspace{4pt}
        \end{subtable}
        \begin{subtable}{\columnwidth}
            \renewcommand{\arraystretch}{1.0}
            \subcaption{\textbf{(50.0m, 50.0m, 2.0m/px $\rightarrow$ 0.5m/px).}}
            \centering
            \vspace{-6pt}
            \begin{tabular}{c
                >{\centering\arraybackslash}p{0.06\textwidth}
                >{\centering\arraybackslash}p{0.08\textwidth}
                >{\centering\arraybackslash}p{0.08\textwidth}
                >{\centering\arraybackslash}p{0.08\textwidth}
                |>{\centering\arraybackslash}p{0.08\textwidth}
                >{\centering\arraybackslash}p{0.08\textwidth}
            }
            \hline
            \multicolumn{5}{c|}{Measure Coverage} & Mem. & Latency \\
            BEV Encoder & Neck & Restore & Decoder & Cache & (GB) & (ms) \\
            \hline
            \ding{51} &  &  &  &  & 2.1 & 134 \\
            \ding{51} & \ding{51} &  &  &  & 2.1 & 142 \\
            \ding{51} & \ding{51} & \ding{51} &  &  & 2.5 & 155 \\
            \ding{51} & \ding{51} & \ding{51} & \ding{51} &  & 2.7 & 164 \\
            \ding{51} & \ding{51} & \ding{51} & \ding{51} & \ding{51} & 3.8 & 164 \\
            \hline
            \end{tabular}
            \vspace{4pt}
        \end{subtable}
        \begin{subtable}{\columnwidth}
            \renewcommand{\arraystretch}{1.0}
            \subcaption{\textbf{(50.0m, 50.0m, 4.0m/px $\rightarrow$ 0.5m/px).}}
            \centering
            \vspace{-6pt}
            \begin{tabular}{c
                >{\centering\arraybackslash}p{0.06\textwidth}
                >{\centering\arraybackslash}p{0.08\textwidth}
                >{\centering\arraybackslash}p{0.08\textwidth}
                >{\centering\arraybackslash}p{0.08\textwidth}
                |>{\centering\arraybackslash}p{0.08\textwidth}
                >{\centering\arraybackslash}p{0.08\textwidth}
            }
            \hline
            \multicolumn{5}{c|}{Measure Coverage} & Mem. & Latency \\
            BEV Encoder & Neck & Restore & Decoder & Cache & (GB) & (ms) \\
            \hline
            \ding{51} &  &  &  &  & 2.1 & 129 \\
            \ding{51} & \ding{51} &  &  &  & 2.1 & 133 \\
            \ding{51} & \ding{51} & \ding{51} &  &  & 2.4 & 144 \\
            \ding{51} & \ding{51} & \ding{51} & \ding{51} &  & 2.6 & 153 \\
            \ding{51} & \ding{51} & \ding{51} & \ding{51} & \ding{51} & 3.7 & 153 \\
            \hline
            \end{tabular}
        \end{subtable}
    }
    \vspace{-18pt}
\end{table}

\noindent\textbf{Efficiency.} In \cref{subfig:latency} and \cref{subfig:memory}, we show the efficiency of our mechanism across increasingly heavier architectural configurations, illustrating how BEVRestore maintains efficiency even as the system architecture grows. Specifically, we add multi-head self-attention (MSA) layers, a well-known operator requiring large memory space, to the fusion neck of BEVFusion and BEVRestore. In \cref{subfig:latency}, BEVFusion exhibits increasing latency, and \cref{subfig:memory} shows diverging memory. In contrast, our model maintains stable cost consumption even when incorporating 32 MSA layers into its architecture. This property will contribute to applying heavy training tricks like vision foundation models. In \cref{tab:footprint}, we investigate the cost footprints of our BEVRestore models and baseline. ``Cache" denotes cache memory of tensors excluding models' consumption. ``BEV Encoder" consists of modality backbones, encoder FPN, and a view transformer, ``Neck" includes fusion and the FPN of the fused BEV feature, ``Restore" is our BEVRestore network, and ``Decoder" means CNN networks for decoding into semantic maps. As shown in the table, although our proposed mechanism requires additional memory for BEVRestore, its utilization of LR BEV contributes to the low cost of the BEV encoder and the neck module's BEV enhancement.

\begin{figure*}[t]
    \vspace{3mm}
    \centering
    \begin{subfigure}[b]{\linewidth}
        \centering
        \includegraphics[width=0.94\linewidth]{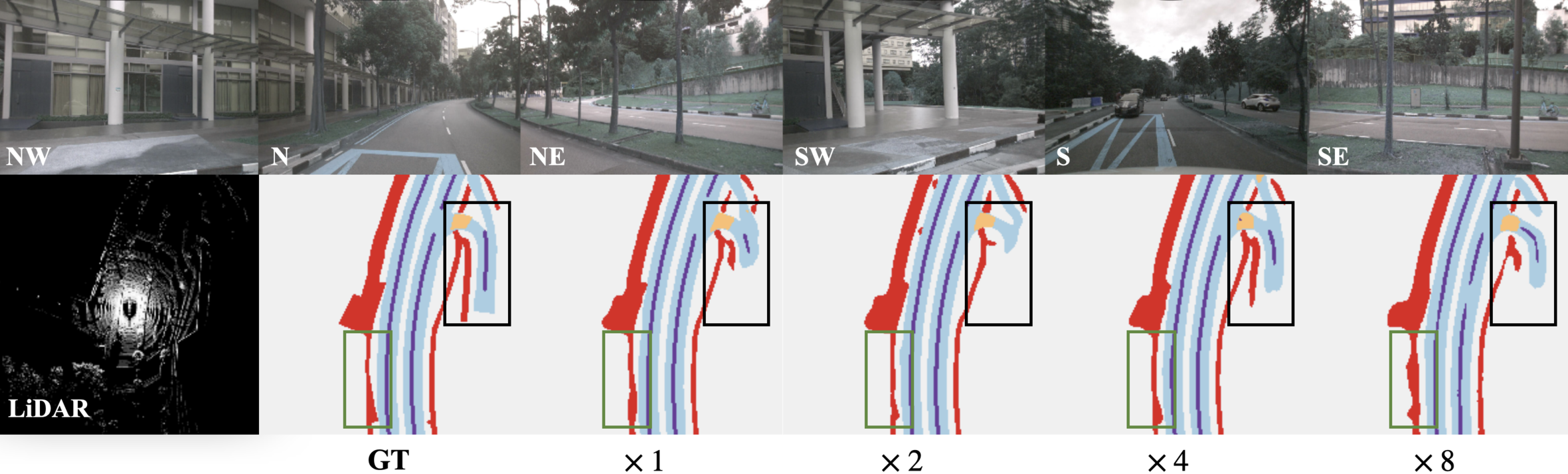} 
    \end{subfigure}
    \caption{\textbf{Restoration Comparisons on varying up-sampling factors.} We use a (-50m, 50m) BEV scope with 0.5m/px, 1.0m/px, 2.0m/px, 4.0m/px LR BEV resolutions for $\times 1, \times 2, \times 4$, and $\times 8$ models, respectively. Their HR BEV space utilizes 0.5m/px HR resolution. The $\times 1$ and $\times 2$ BEVRestores fail to learn high-level scene features (Black ROIs). In contrast, $\times 8$ BEVRestore suffers from inaccurate prediction (Green ROIs) caused by severe loss of prior, compared to the $\times 4$ BEVRestore.}
    \label{fig:abl}
    \vspace{-5pt}
\end{figure*}

\begin{figure*}[t]
    \centering
    \begin{subfigure}[b]{\linewidth}
        \centering
        \includegraphics[width=0.185\textwidth]{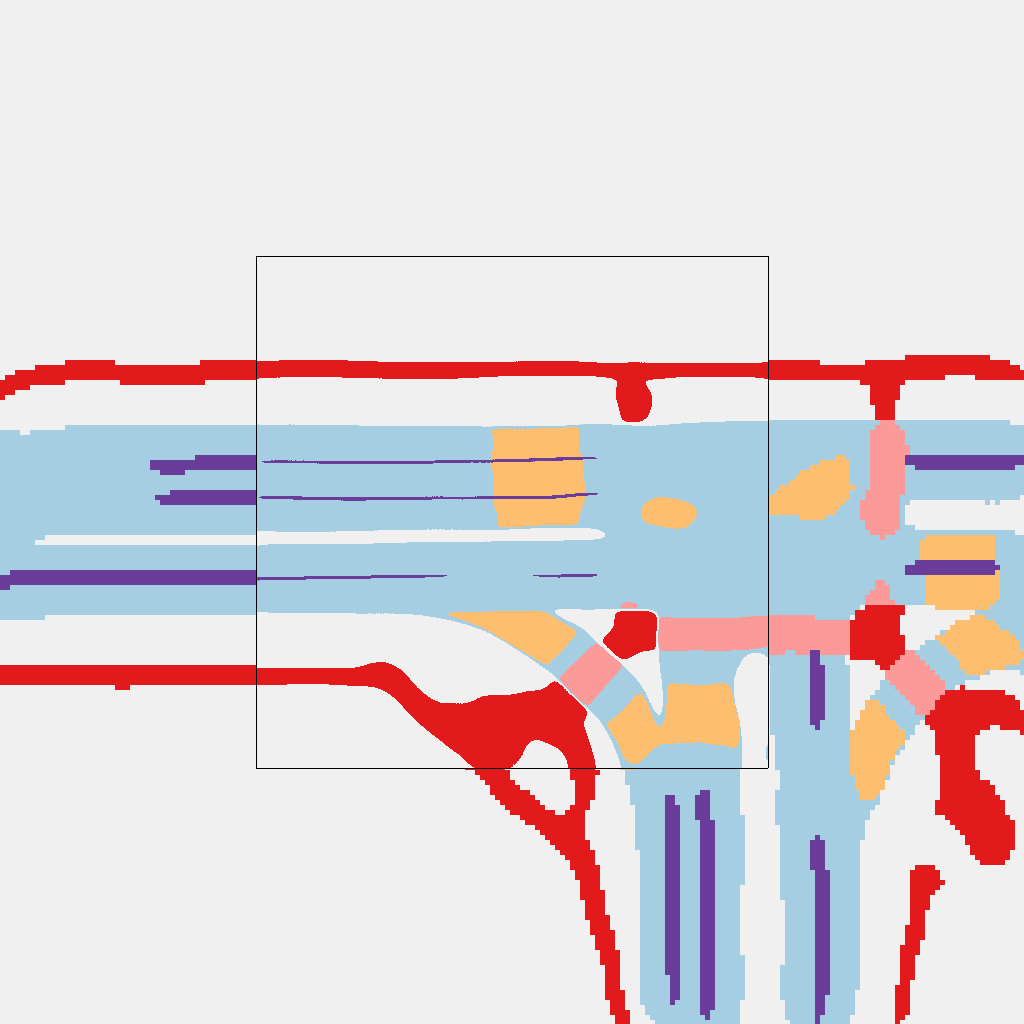}
        \includegraphics[width=0.185\textwidth]{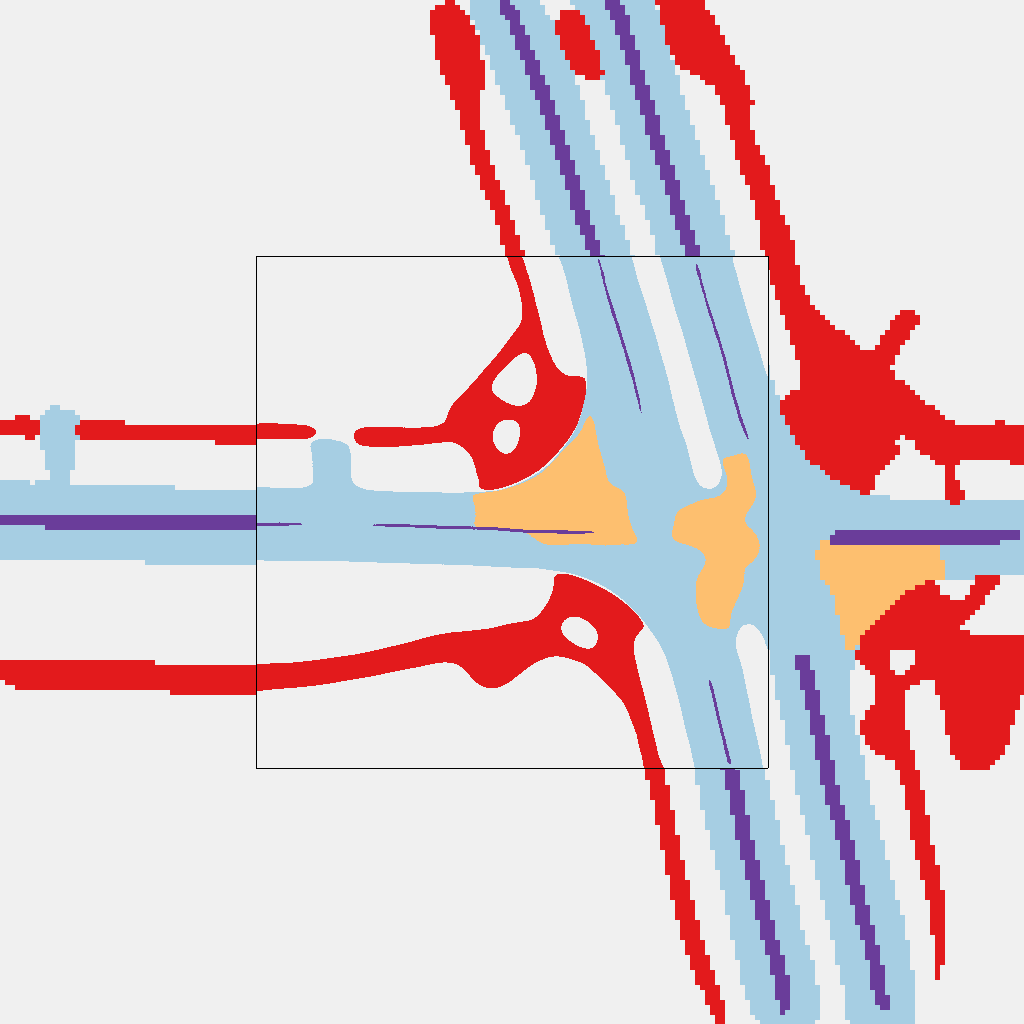}
        \includegraphics[width=0.185\textwidth]{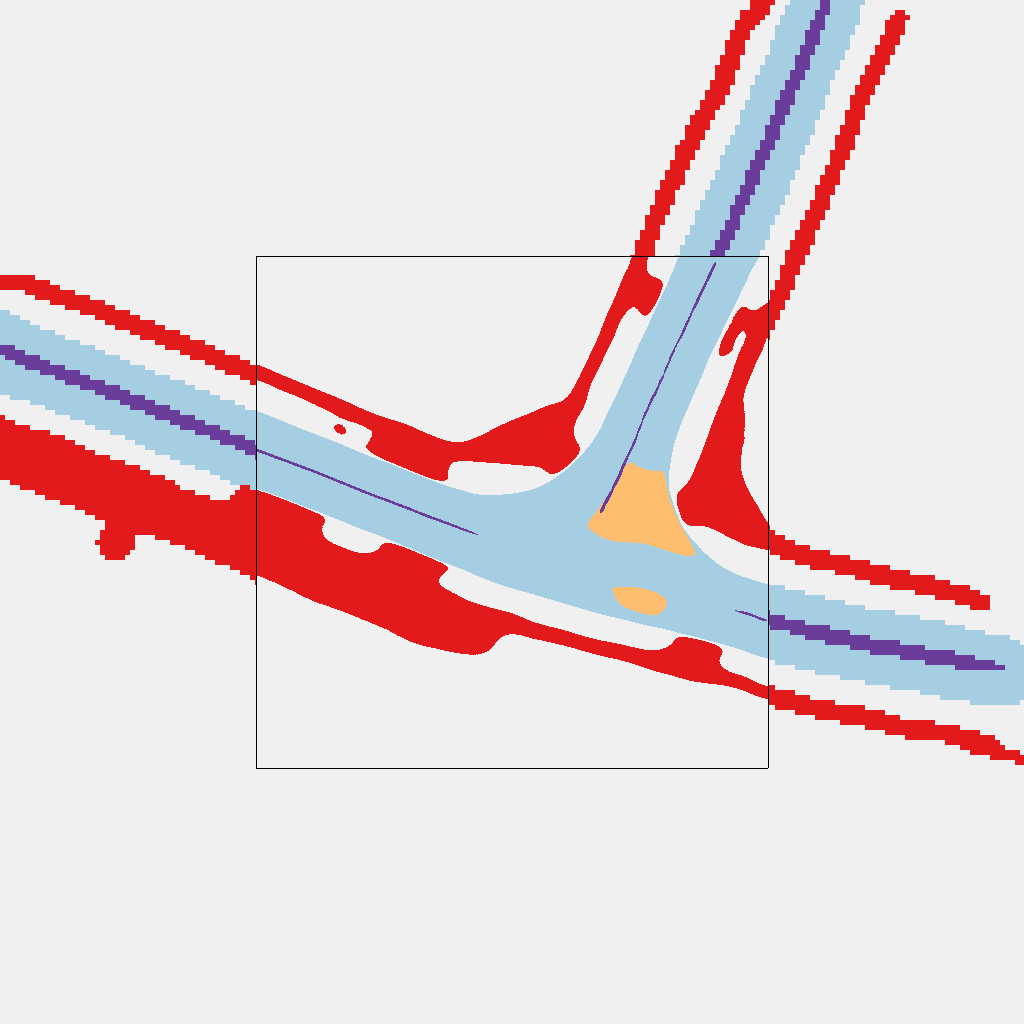}
        \includegraphics[width=0.185\textwidth]{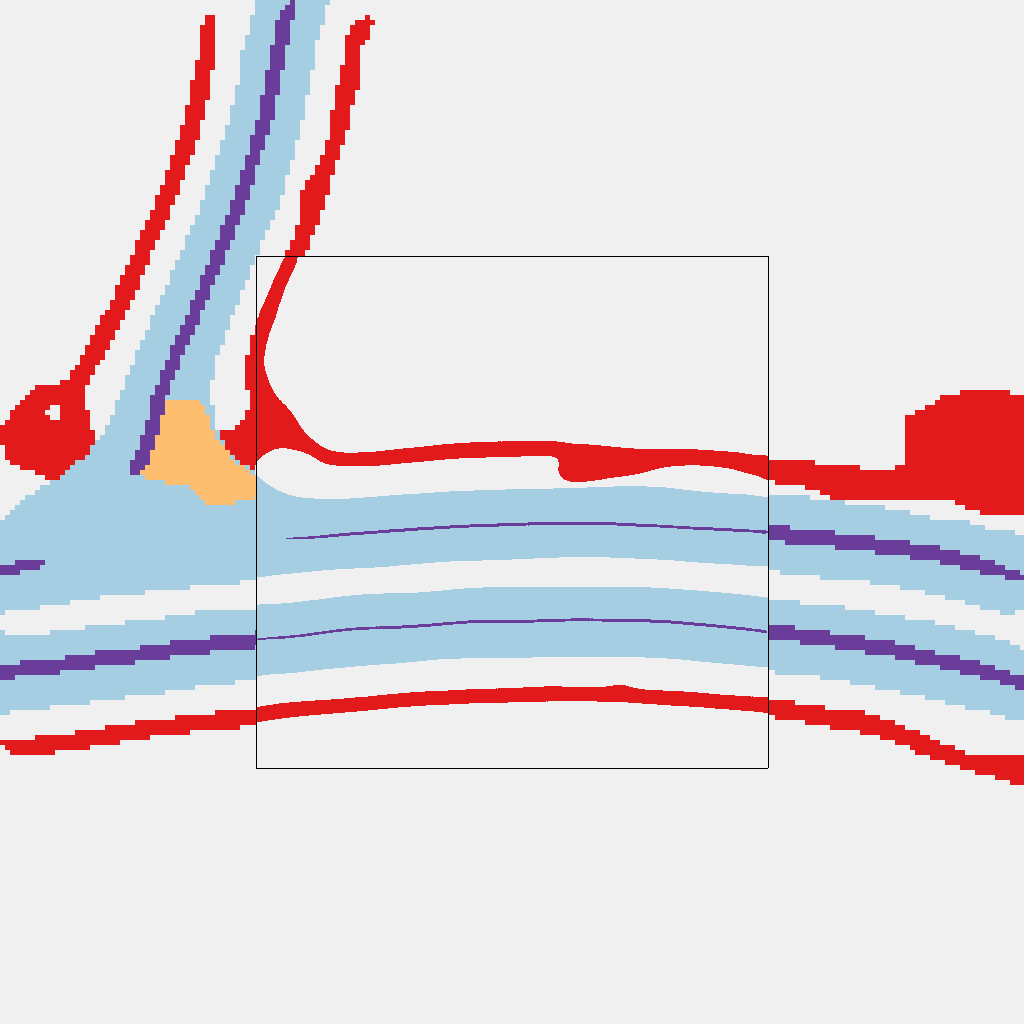}
        \includegraphics[width=0.185\textwidth]{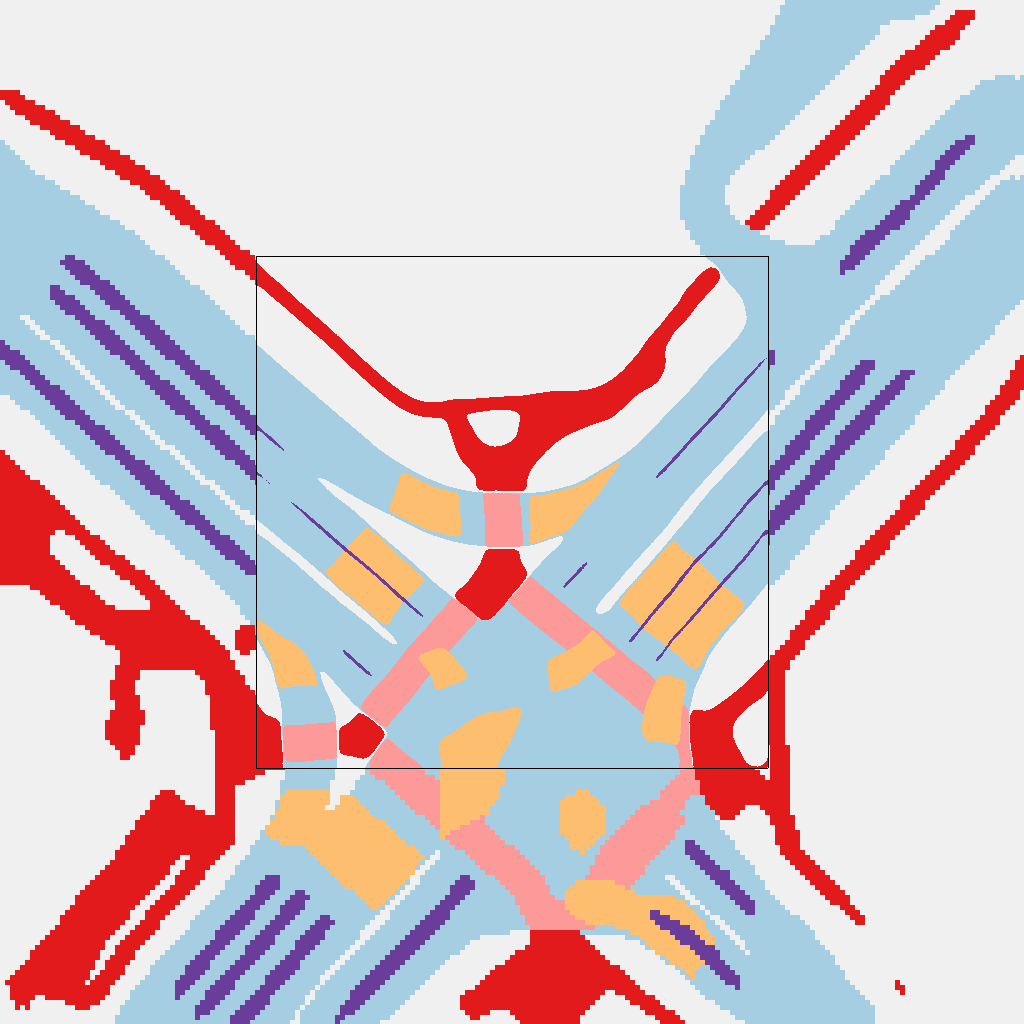}
    \end{subfigure}
    \caption{\textbf{HR perception using $\times 8$ BEVRestore.} Maps have 0.1m/px inner and 0.8m/px outer scopes.}
    \label{fig:doublebell}
    \vspace{-5pt}
\end{figure*}

\begin{figure*}[t]
    \centering
    \begin{subfigure}[b]{0.95\linewidth}
        \centering
        \includegraphics[width=\textwidth]{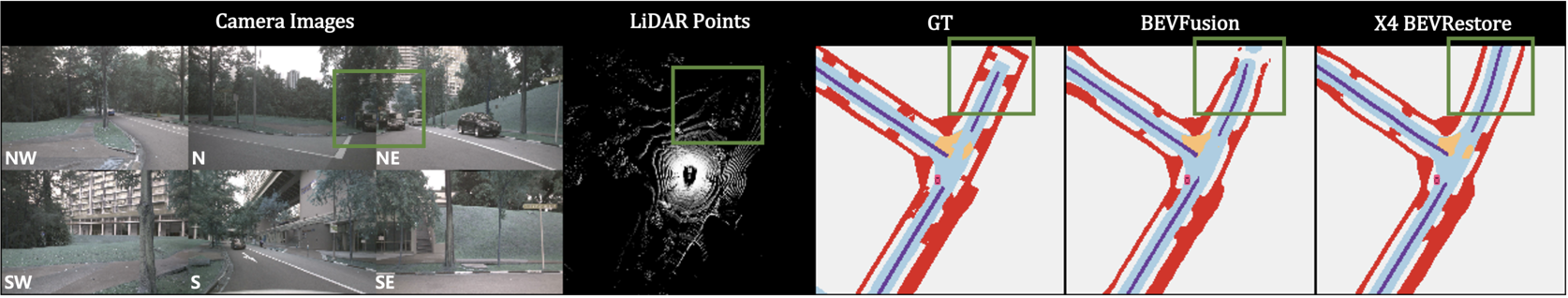} 
    \end{subfigure}
    \caption{\textbf{Overconfident Issues.} \textcolor{OliveGreen}{Green ROIs}: Far regions occluded by cars. Our map construction struggles with the overconfident prediction of uncertain regions.}
    \label{fig:failure}
    \vspace{-16pt}
\end{figure*}

\subsection{Qualitative Results}
\noindent\textbf{BEVRestore.} In \cref{fig:qual}, we compare up-sampling methods qualitatively to demonstrate the limited map construction of the hand-crafted approaches (nearest neighbor, bilinear, bicubic interpolations). The images of the first row are camera images, and the first column of the second row shows LiDAR points of BEV. The hand-crafted methods fail to reconstruct blocky LR BEV showing uncompleted construction of deteriorated class labels like walkways, and road lanes. In addition, they fail to restore aliasing as in the edges of ``walkway" labels. However, our pixel shuffle-based learnable upsampler resolves the problems showing successful restorations of the aliased and blocky BEV features.


\noindent\textbf{Up-scaling Factor.} In \cref{fig:abl}, we explore various up-scale factors, qualitatively. BEVRestores with low up-scale factors ($\times 1, \times 2$) fail to capture high-level scene features but preserve local details as shown in the black and green ROIs, respectively. In contrast, BEVRestores with high up-scale factors ($\times 4, \times 8$) facilitate approximating the high-level description. However, excessive BEVRestore with up-scale factors like $\times 8$ cause severe signal aliasing and blocky artifacts, thereby imposing a reasonable selection of a specific point that has compactness and tolerant signal quality.

\noindent\textbf{HR BEV Perception.} In \cref{fig:doublebell}, we demonstrate our mechanism's HR perception. The task severely limits BEVFusion to training its model due to the slow training speed and huge backpropagation memory. However, the model with our mechanism is free from the diverging training costs of learning HR BEV representation as our $\times 8$ BEVRestore model, which respectively adopts 0.8m/px and 0.1m/px for LR and HR BEV resolution, has efficient training pipelines. Our proposed mechanism enables efficient training, facilitates HR perception, and enhances map construction performance.

\noindent\textbf{Overconfident Issue.}
\label{sec:failure}
We have introduced an effective strategy for capturing the structured patterns of the urban environments \cite{lynch1964image} in LR BEV space and up-sampling the features to HR BEV space. However, this method can sometimes yield overconfidence in the predicted semantic labels for occluded regions, leading to errors as shown in \cref{fig:failure}. We believe that incorporating temporal information for enriching prior could enhance the delineation of uncertain areas in future in-depth studies.

\section{CONCLUSIONS} \label{sec:conclusion}
We suggested the BEVRestore, a novel mechanism tailored with a BEVRestore, aimed at a memory-efficient precise construction of bird's eye view (BEV) maps. BEVRestore resolved diverging training costs, thereby facilitating the training for the extraction of high-resolution BEV features. Our experiments in BEV map construction demonstrated that our proposed approach finely, and precisely restores BEV features, achieving 4.3\%, 10.1\%, and 8.2\% mIoU gains on BEV segmentation with camera, LiDAR, and LiDAR-camera fusion, respectively, over the existing baseline model BEVFusion. Moreover, the adaptability and effectiveness of the BEVRestore mechanism for vectorized map construction are validated respectively showing the mIoU gains of 9.1\% mAP and 2.2\% mIoU for vectorized and rasterized map construction of HDMapNet and 1.2\% mIoU gain for rasterized map construction of BEVFormer. The experimental results show the solid compatibility and memory efficiency of BEVRestore, establishing a new baseline for precise BEV map construction which is essential for safe autonomous driving.

\bibliographystyle{IEEEfull}
\bibliography{IEEEfull}

\end{document}